\documentclass{CVM}

\CVMsetup{
type      = {Research/Review Article},
doi       = {CVM.XXXX},
title     = {Multi-level Dynamic Style Transfer for NeRFs},
author    = {Zesheng Li$^{1}$, Shuaibo Li$^{2}$, Wei Ma$^{1}$\cor{}, Jianwei Guo$^{3}$, and Hongbin Zha$^{4}$\\
},
runauthor = {Z. Li, S. Li, W. Ma, J. Guo, H. Zha},
abstract  = {
As the application of neural radiance fields (NeRFs) in various 3D vision tasks continues to expand, numerous NeRF-based style transfer techniques have been developed. However, existing methods typically integrate style statistics into the original NeRF pipeline, often leading to suboptimal results in both content preservation and artistic stylization. In this paper, we present multi-level dynamic style transfer for NeRFs (MDS-NeRF), a novel approach that reengineers the NeRF pipeline specifically for stylization and incorporates an innovative dynamic style injection module.  Particularly, we propose a multi-level feature adaptor that helps generate a multi-level feature grid representation from the content radiance field, effectively capturing the multi-scale spatial structure of the scene. In addition, we present a dynamic style injection module that learns to extract relevant style features and adaptively integrates them into the content patterns. The stylized multi-level features are then transformed into the final stylized view through our proposed multi-level cascade decoder. Furthermore, we extend our 3D style transfer method to support omni-view style transfer using 3D style references. Extensive experiments demonstrate that MDS-NeRF achieves outstanding performance for 3D style transfer, preserving multi-scale spatial structures while effectively transferring stylistic characteristics.
},
keywords  = {Style Transfer, NeRF, 3D, Multi-level  Features},
copyright = {The Author(s)},
}

\begin{document}

\maketitle

\enlargethispage{-3pt}
\begin{figure}[b] \vskip -4mm
\small\renewcommand\arraystretch{1.3}
\begin{tabular}{p{80.5mm}} \toprule\\ \end{tabular}
\vskip -4.5mm \noindent \setlength{\tabcolsep}{1pt}
\begin{tabular}{p{3.5mm}p{80mm}}
$1\quad $ &  Beijing University of
Technology, China. E-mail: S202274160@emails.bjut.edu.cn;
mawei@bjut.edu.cn.\\
$2\quad $ & The Hong Kong University of Science and Technology (Guangzhou), China. E-mail:
sli270@connect.hkust-gz.edu.cn.\\
$3\quad $ & School of Artificial Intelligence, Beijing Normal University, and MAIS, Institute of Automation, Chinese Academy of Sciences, China. E-mail: jianwei.guo@bnu.edu.cn.\\
$4\quad $ & Key Laboratory of Machine Perception (MOE), School of IST, Peking University, China. E-mail: zha@cis.pku.edu.cn.\\

\end{tabular} \vspace {-3mm}
\end{figure}

\section{Introduction}
3D style transfer aims to apply the stylistic characteristics of a reference source to a 3D scene while preserving the spatial structures of the target. This technique, which bridges artificial intelligence and 3D artistic creation, has diverse applications in augmented reality (AR), virtual reality (VR), and home interior design. Recent advances in 3D scene representation \cite{3,4,5} and style transfer mechanisms \cite{7,8,chen2021nips,11,12} have propelled the field of 3D stylization \cite{fan2022unified,Mu_2022_CVPR,UPST}. In this paper, we focus on  style transfer using neural radiance fields (NeRFs) \cite{1}, which has recently gained increasing attention due to its advantages of high-quality rendering and storage efficiency.

NeRF-based 3D style transfer methods can be categorized into two types. The first directly optimizes NeRFs to match the reference style \cite{NeRFArt,IN2N,vica-nerf}. While these methods effectively apply stylistic characteristics, they require NeRF optimization for each individual style reference, which limits their practical applicability. The second  conducts stylization during the rendering stage (e.g., StylizedNeRF~\cite{Huang_2022_CVPR}, StyleRF~\cite{Liu_2023_CVPR}, and Hyper~\cite{Chiang_2022_WACV}), enabling multiple styles or even zero-shot style transfer once the models have been trained. However, these methods are constrained in their ability to encode scene features, resulting in either inadequate stylization or compromised spatial structures in their results.

To address these challenges, we propose a novel framework for zero-shot NeRF style transfer, multi-level dynamic style transfer for NeRFs (MDS-NeRF). MDS-NeRF represents the 3D content as a multi-level feature grid using an innovative multi-level feature adaptor. It facilitates the integration of diverse and scale-appropriate style characteristics while preserving the spatial structures of the target. Once the features have been adjusted via style injection, they are decoded separately and then gradually fused using our proposed multi-level cascade decoder. To mitigate inconsistent results between views that may arise from upsampling during decoding, we render the multi-level features at the same resolution as the original view.

To perform style injection into the rendered multi-level features, a widely adopted approach known as AdaIN \cite{6} substitutes their statistics with those of the multi-scale style features extracted by convolutional neural networks (CNNs). However, in our method, while the content features are derived from the radiance field through the proposed feature adaptor (trained with the supervision of upsampled VGG \cite{simonyan2014very} features), the style features are extracted directly using VGG. This discrepancy results in differences between the content and style feature patterns. To address this issue and effectively aggregate appropriate stylistic characteristics into the multi-level content representation, we propose a dynamic style injection (DSI) module, inspired by DIN \cite{9}. This DSI learns to extract and infuse stylistic characteristics by referencing both spatial and cross-channel contexts present in the style features. This helps generate view-consistent results with rich style characteristics and well-preserved spatial structures. Furthermore, leveraging the multi-level feature grid along with DSI facilitates 3D style mixing by integrating individual styles into different levels of the content features.

Moreover, existing 3D style transfer methods that primarily use 2D images as style references cannot transfer omni-view styles from 3D references. However, with rapid advances in 3D reconstruction and generation, 3D references have become increasingly accessible. In our proposed MDS-NeRF framework, we introduce a multi-view style injection mechanism that enables both effective 2D-to-3D and 3D-to-3D style transfer.

In summary, we claim the following contributions:
\begin{itemize}
\item The \emph{MDS-NeRF framework}  for style transfer for NeRFs. It features an innovative NeRF rendering pipeline tailored specifically for stylization, along with a dynamic style injection mechanism. MDS-NeRF demonstrates superior performance to existing methods in terms of content preservation and artistic stylization.

\item A \emph{multi-level feature adaptor} which represents the content radiance field as a multi-level feature grid, paired with a multi-level cascade decoder that transforms the stylized feature grid into a rendered view. This approach facilitates the injection of diverse and scale-appropriate style characteristics while preserving the spatial structures of each view.

\item A \emph{dynamic style injection module} which  adaptively learns the preferred style characteristics for each level of the content feature grid, addressing  pattern discrepancies between the content and style features. It demonstrates superior performance  to the widely used AdaIN style injection method.

\item We extend MDS-NeRF to enable \emph{3D-to-3D omni-view style transfer} through multi-view aligned style injection, allowing our method to effectively leverage widely accessible 3D references.
\end{itemize}

\begin{figure*}[t!]
    \centering
    \includegraphics[width=\linewidth]{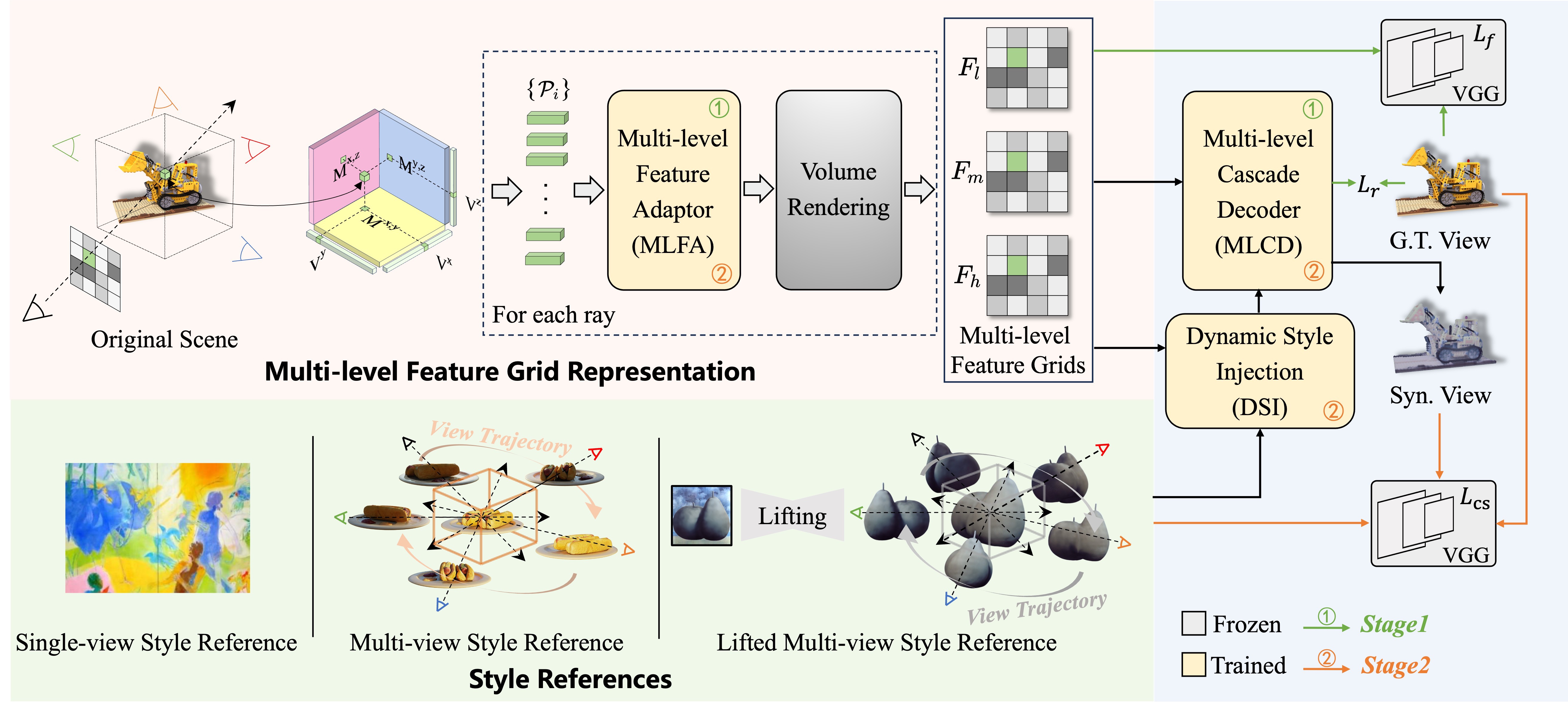}
    \caption{\textbf{Overview of MDS-NeRF.} MDS-NeRF features a multi-level feature grid representation along with a dynamic style injection (DSI) module. For a given view to be rendered, we map the basis point features $\{\mathcal{P}_i\}$ sampled along each ray into multi-level pixel features using the multi-level feature adaptor (MLFA) and volume rendering, obtaining the multi-level content features, denoted  $F^\ell,{\ell}\in\{{\ell}_l,{\ell}_m,{\ell}_h\}$. Given a style reference, which could be in either 2D or 3D form, the proposed DSI selects the preferred style characteristics and injects them into the content features. Finally, the altered multi-level content features are decoded by the multi-level cascade decoder (MLCD) to produce a stylized RGB view. \emph{Stage1} and \emph{Stage2} refer to the two training stages: the multi-level feature grid reconstruction stage and the  stylization stage, respectively.}
    \label{fig:pipeline}
\end{figure*}

\section{Related Work}

\subsection{3D style transfer}

Many methods for 3D style transfer have been developed specifically for explicit 3D representations such as meshes, point clouds, and volumes. For example, StyleMesh \cite{stylemesh} transfers styles to 3D meshes of indoor scenes by leveraging depth information and surface normals to achieve view-independent stylization. LSNV \cite{huang2021learning} consistently stylizes novel views of a 3D scene through point cloud aggregation and linear feature transformation. \cite{efficientnst} introduced a suite of upgrades to volumetric neural style transfer that result in faster, simpler, more controllable stylization with reduced artifacts.

Recently, a variety of methods have been proposed for 3D style transfer based on neural radiance fields. Given an image as a style reference, ARF \cite{intro_ARF} optimizes the NeRF representation via a content loss combined with a style loss defined via nearest neighbor feature matching. SNeRF \cite{nguyen2022snerf} uses an optimization strategy that gradually achieves 3D-aware stylization by utilizing existing 2D style transfer methods. Ref-NPR  \cite{Ref-NPR} stylizes a single 2D view of a pre-trained radiance field and uses it to generate stylized novel views through a ray registration process, followed by occlusion filling based on semantic correspondence. Optimization-based style transfer has also been applied to stylize 3D Gaussian splatting (3DGS). For example, G-Style \cite{GStyle} repeatedly performs stylization and geometry matching of the content scenes to achieve stylized Gaussian splatting based on a style reference image. StylizedGS \cite{StylizedGS} applies color histogram matching with the reference style  and then uses filter-based 3DGS refinement to address artifacts. Multiple losses, including depth preservation loss and stylization losses, are incorporated to facilitate style transfer while preserving content geometric attributes. While these methods produce effective stylization results, they are time-consuming due to the necessity of performing optimization for each individual reference style.

Apart from  optimization-based NeRF stylization methods, several methods have been developed to achieve multi-style transfer. Hyper leverages hypernetwork \cite{Ha2016hyper} to compute the parameters of a color multi-layer perceptron (MLP) with a given style reference, while StylizedNeRF learns a style module to alternate the color of queried 3D points according to the latent codes of style features. Both methods are limited by the reconstruction accuracy of NeRFs, leading to blurry rendering results.  StyleRF performs zero-shot 3D style transfer on the rendered 2D features, allowing the production of detailed results using an arbitrary style reference. However, it only transfers style to high-level features, producing noisy artifacts in its stylized results. In contrast, our proposed MDS-NeRF addresses these limitations by extracting preferred style characteristics from multiple levels of style features and adaptively injecting them into the multi-level content features. Our approach achieves effective stylization while preserving  spatial structures within target scenes.

Moreover, existing NeRF-based style transfer methods generally use 2D images as style references. In this paper, in addition to 2D-to-3D style transfer, we extend our approach to facilitate 3D-to-3D style transfer. This enhancement allows our method to utilize widely accessible 3D references and to achieve omni-view style transfer.

\subsection{3D scene NeRF representation}
3D representations have been widely studied in the past, with common forms including point clouds~\cite{Rente2019TMM}, volumes~\cite{3dprinting}, meshes, etc.
Recently introduced, NeRFs (neural radiance fields) allow a novel approach to view synthesis  that achieves realistic rendering. An implicit function is used to encode  color and density information for each point within the scene, offering advantages in terms of spatial representation continuity and efficiency. However, the training speed of the NeRF pipeline is relatively slow, which poses challenges for practical applications.

Several methods have introduced explicit representations to accelerate  NeRF training while improving both the appearance and spatial structures of reconstructed scenes. Specifically,
\cite{sun2022direct, Fridovich-Keil_2022_CVPR, Hou2024nerf} introduced voxel grids, while \cite{muller2022instant} leveraged hash tables to facilitate faster value queries.
Additionally, TensorRF \cite{2}, which models a scene as a low-rank representation based on tensor decomposition, achieves high reconstruction quality with reduced memory costs. Instead of directly representing the color and density values, some works~\cite{Niemeyer_2021_CVPR,NEURIPS2020_b4b75896} explore feature representations aimed at further alleviating the complexity associated with scene reconstruction.

Although significant progress has been made in NeRF representations, current methods often overlook the importance of adequately encoding spatial structures during rendering, which is crucial for effective 3D style transfer. In this work, we present a new NeRF pipeline by introducing a multi-level feature grid representation. The grid representation effectively reveals the multi-scale spatial structure of the content, facilitating high-quality 3D style transfer. Furthermore, with the multi-level feature grid representation, our method can achieve style mixing, allowing for more creative style editing.

\section{Overview}\label{sec:Overview}

Given a 3D scene in the form of a pre-trained basic NeRF representation and a style reference, the objective of this work is to inject the style characteristics into the NeRF rendering pipeline to generate stylized views. Fig.~\ref{fig:pipeline} illustrates the proposed MDS-NeRF framework. For each view, the queried feature points along each ray are transformed into multi-level feature points via the proposed MLFA module (see Sec. \ref{sub:sec3.1}). All  feature points from all rays form a  multi-level feature grid for that view through volume rendering. We then extract multi-level style features using VGG. The multi-level content features, along with the style features, are sent to the dynamic style injection module (see Sec.~\ref{sub:sec3.2}), where the style information is selected and injected into the corresponding content features. Finally, the modified content features are decoded by the multi-level cascade decoder (MLCD) to obtain the stylized result for that particular view. Additionally, the proposed MDS-NeRF is designed to accommodate both 2D images and 3D models as references (see Sec.~\ref{sub:sec3.3})  for stylization. The training of MDS-NeRF occurs in two stages. The first, referred to as the multi-level feature grid reconstruction stage, involves training the MLFA and MLCD modules to enable  reconstruction of features  aligned with the ground-truth view. In the second stylization stage, we focus on training the DSI and MLCD modules to enable  zero-shot style transfer. Additionally, a few parameters within the MLFA module related to style suppression for the content features are learned in this stage. The two stages are denoted \emph{Stage1} and \emph{Stage2}, respectively.

\section{Method}\label{sec:Method}

\subsection{Multi-level Feature Grid Representation}\label{sub:sec3.1}

We propose a multi-level feature grid representation to reveal the multi-scale spatial structures of the content observed in the view. This approach facilitates intricate content-style matched stylization, thereby integrating rich stylistic characteristics while preserving the spatial structure of the content.  Although previous methods,  such as StyleRF, have explored feature grid representations for zero-shot style transfer, they typically employ only a single-level grid. Such designs  often result in artifacts in regions with complex textures and spatial structures, as shown later in Fig.~\ref{fig:qualitative compare1}.

The advantages of the proposed multi-level feature grid representation are as follows. First, it enables style injection at multiple feature levels, thereby allowing for the integration of more comprehensive style characteristics for effective stylization. Second, the stylized result can be obtained from the multi-level stylized features, which is crucial for retaining the injected style effects, particularly with respect to fine details. Third, injecting preferred styles into each individual scale of the content promotes the maintenance of multi-scale spatial structures after stylization.

\begin{figure}[t!]
    \centering
    \includegraphics[width=\linewidth]{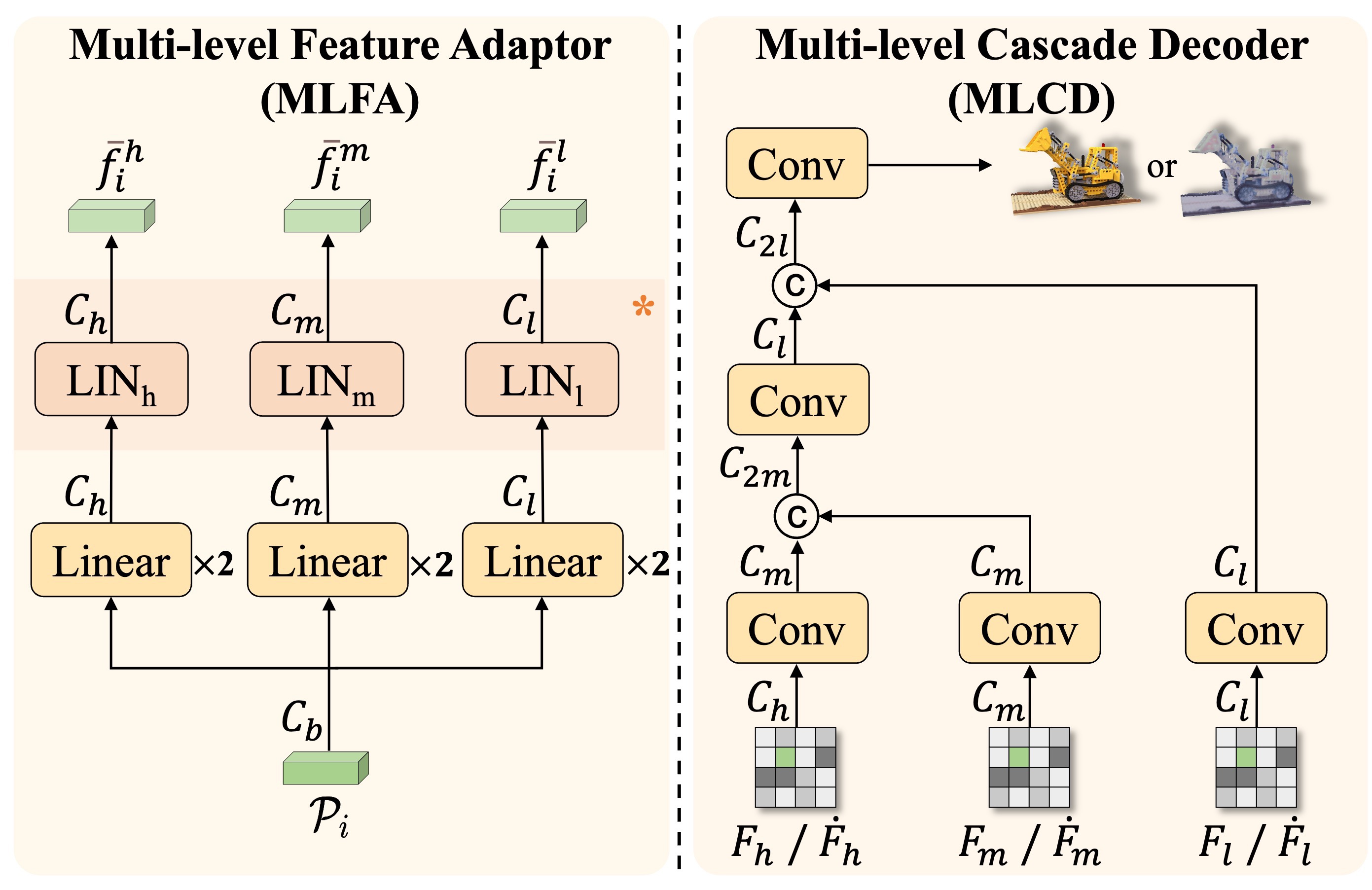}
    \caption{\textbf{MLFA and MLCD.}
    The multi-level feature adaptor (MLFA) transforms basic point features into multi-level point features and employs learnable instance normalization (LIN) to suppress the styles in the content features.  LIN (indicated with *) is applied during stylization. The multi-level cascade decoder (MLCD) decodes the multi-level content features in the first training stage, or the stylized content features during stylization, into an RGB image in a cascaded manner, progressively integrating information from different feature levels.}
    \label{fig:MLFA and MLCD}
\end{figure}

While modifying basic NeRF representations for multi-level scene expression is complex and inefficient, we propose the MLFA module (illustrated in Fig.~\ref{fig:MLFA and MLCD}) to effectively encode multi-level features of NeRF scenes. MLFA can be applied to various NeRF representations. In our implementation, we utilize a set of low-rank matrices $M$ and vectors $V$ as the base NeRF representation, following the approach of TensoRF. Given a point $i$, $i \in \left [ 1,\dots,N \right ]$, located on a ray $r$ from  viewpoint $v$, the linear layers in MLFA adaptively transfer the basic point feature $\mathcal{P}_i$, $\mathcal{P}_i \in \mathbb{R}^{C_b}$, sampled from the $MV$s, into multi-level point features, denoted  $\bar{f}_i^{{\ell}}$, $\bar{f}_i^{{\ell}}\in \mathbb{R}^{C_{{\ell}}}$. Here, ${{\ell}}\in\{{{\ell}}_{l},{{\ell}}_{m},{{\ell}}_{h}\}$, representing  low, middle or high level  features. $C_b$ and $C_{{\ell}}$ denote the numbers of channels  of the basic point features and  of the transformed point features at level ${\ell}$, respectively.

As illustrated in Fig.~\ref{fig:MLFA and MLCD}, we use a learnable instance normalization (LIN) operation in MLFA to remove the style characteristics from each level of the content features before performing the subsequent dynamic style injection (see Sec. \ref{sub:sec3.2}). Note that the LIN operations are applied during stylization and skipped in the first training stage for multi-level feature grid reconstruction. Mathematically, the process of obtaining multi-level point features in MLFA can be described as:
\begin{equation}\\
\bar{f}_i^{{\ell}}=\begin{cases}\text{MLP}_{{\ell}}(\mathcal{P}_i), &Stage1\\\text{LIN}_{{\ell}}\big(\text{MLP}_{{\ell}}(\mathcal{P}_i)\big), &Stage2\end{cases}
\label{eq:1}
\end{equation}
\begin{equation}
  \text{LIN}_{{\ell}}(x)=\frac{x-\tilde{\mu}_{{\ell}}}{\tilde{\sigma}_{{\ell}}},
\end{equation}
where $\text{MLP}_{{\ell}}$ denotes the linear layers for transferring the $\mathcal{P}_i$ at feature level ${\ell}$. $\tilde{\mu}_{{\ell}}$ and $\widetilde{\sigma}_{{\ell}}$ are two learnable parameters that are optimized during the training stage for stylization. Unlike traditional instance normalization, LIN can normalize point features uniformly across the entire scene space, theoretically achieving multi-view-consistent results.

Regarding the geometry, following StylizedNeRF, we use the opacity field from the pre-trained scene representation to provide the volume density $\sigma_i$ of the sampled point $i$. After obtaining $\{\bar{f}_i^{{\ell}}\}$ and $\{\sigma_i\}$ for the sampling points on ray $r$, we can calculate the corresponding multi-level features $f_r^{{\ell}} \in \mathbb{R}^{C_{{\ell}}}$ of the pixel through volume rendering, which can be expressed as:
\begin{equation}
    f_r^{{\ell}} = \sum_{i=1}^{N} \omega_i \bar{f}_i^{{\ell}},
    \label{Eq:frl}
\end{equation}
\begin{equation}
    \omega_i=\exp\left(-\sum_{q=1}^{i-1}\sigma_q\Delta_q\right)\left(1-\exp(-\sigma_i\Delta_i)\right) ,
\end{equation}
where $\omega_i$ denotes the weight of the sampled point $i$, and $\Delta_i$ is the distance between contiguous sampled points. By repeating the above process for each ray, we can obtain the multi-level feature maps $F_v^{{\ell}} \in \mathbb{R}^{H\times W\times C_{{\ell}}} $ for the given view $v$, with height and width $H$ and $W$ respectively.

To prevent view inconsistency caused by  upsampling  during decoding, we render the feature maps at the same resolution as that of the original view. We upsample the feature maps $\hat{F}_v^{{\ell}} \in \mathbb{R}^{H_{{\ell}} \times W_{{\ell}} \times C_{{\ell}}}$, which are extracted by the pre-trained VGG, and use them as supervision to optimize the feature grid; $H_{{\ell}}$ and $W_{{\ell}}$ are the height and width of the extracted features at level ${\ell}$, respectively. Mathematically, the loss function $L_f$, is defined as:
\begin{equation}
    L_{f}=\sum_{v\in \mathcal{V}}\sum_{l\in \mathcal{L}}\left\|F_{v}^{{\ell}}-\text{upsamp}_{{\ell}}\big(\hat{F}_{v}^{{\ell}}\big)\right\|_{2}^{2},
    \label{eq:5}
\end{equation}
where $\mathcal{V}$ and $\mathcal{L}$ are the set of training views and feature levels, respectively. $\text{upsamp}_{{\ell}}()$ denotes the upsampling operation that equalizes the size of the extracted feature maps $\hat{F}_v^{{\ell}}$ with that of the rendered feature maps $F_v^{{\ell}}$ at level ${\ell}$.

To decode the multi-level feature grid into an RGB image at view $v$, we draw inspiration from the U-Net architecture utilized in segmentation tasks \cite{unet} to propose the MLCD module. As illustrated in Fig.~\ref{fig:MLFA and MLCD}, MLCD progressively integrates the input features across three levels. During the first training stage, the input features are content features, while during stylization, they are stylized features. Initially, the high-level and middle-level representations are convolved into features with the same number of channels, after which they are concatenated along the channel dimension. The resulting integrated features are further combined with the low-level features through a process of channel adjustment via convolution, followed by concatenation along the channel dimension. Finally, this fused output is processed by a convolution layer to obtain an RGB image $I_v$, $I_v \in \mathbb{R}^{H\times W\times 3}$.

Note that the resolution of the multi-level features matches that of the view image, avoiding the need for upsampling during decoding. This implies that the original decoding convolution operations used in U-Net are no longer suitable, as the receptive fields of the convolution kernels have been altered. To solve this problem, we utilize dilated convolution to appropriately expand the receptive fields, ensuring their alignment with those in the corresponding layers of the VGG model. The dilation rate is increased when convolving over higher level features. Additionally, to reduce the learning cost while maintaining the appearances and structures of the rendered images, MLCD is trained in both the multi-level feature grid construction stage and the stylization stage. The total optimization loss $L_g$ for the multi-level feature grid combines $L_f$ and an RGB recovery loss $L_r$:
\begin{equation}
    L_g = L_f + L_r ,
\end{equation}
\begin{equation}
    L_{r}=\sum_{v\in \mathcal{V}}\left\|\hat{I}_{v}-I_{v}\right\|_{2}^{2} ,
\end{equation}
where $\hat{I}_v$ and $I_v$ denote the ground truth image and the decoded image from  observing view $v$, respectively.

\subsection{Dynamic Style Injection}\label{sub:sec3.2}

\begin{figure}[t!]
    \centering
    \includegraphics[width=\linewidth]{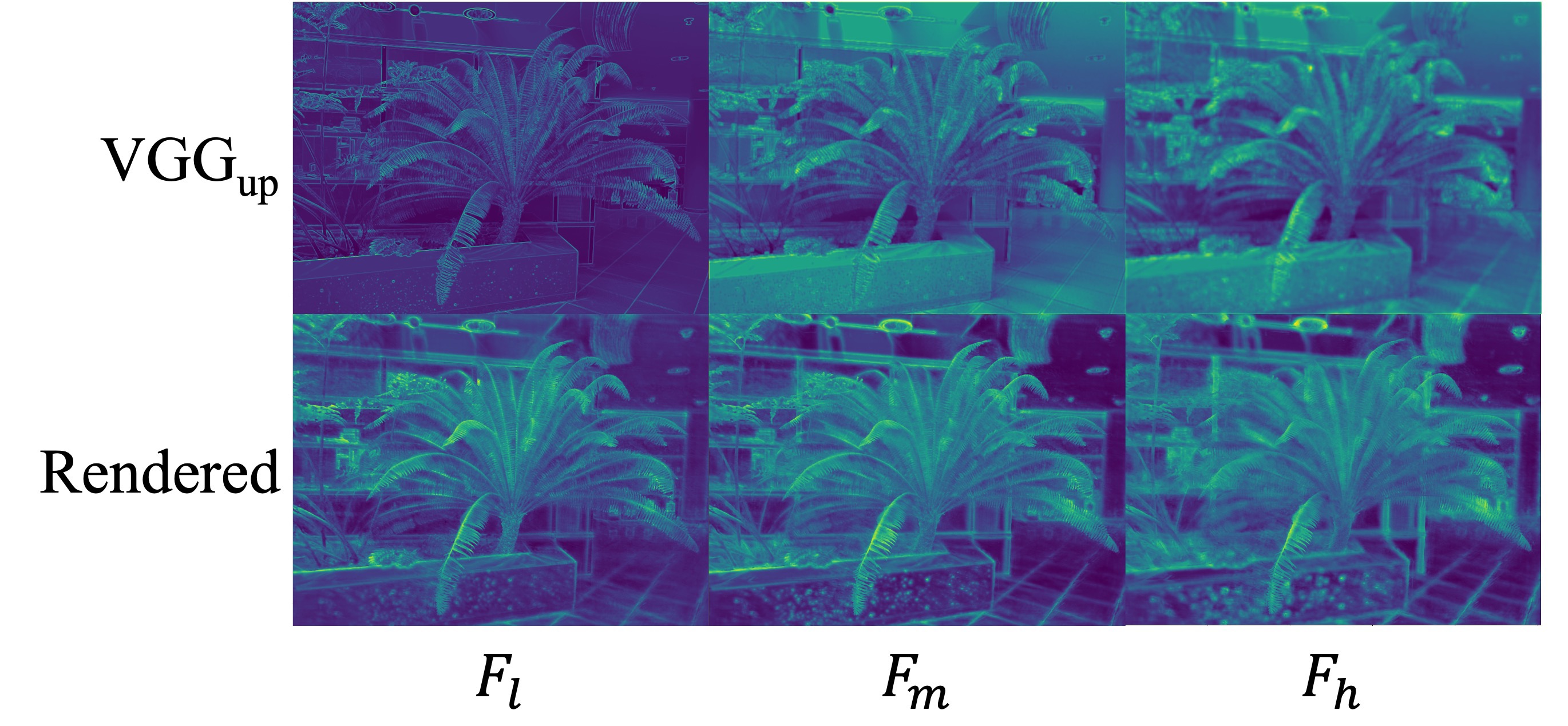}
    \caption{\textbf{NeRF rendered features and VGG extracted features.} The rendered features exhibit discrepancies compared to the VGG extracted features.}
    \label{fig:feature comparison}
\end{figure}

Having obtained the multi-level feature grid representation of the 3D content, we can inject the reference style into these  features. A naïve style injection method is AdaIN, which directly extracts the feature statistics of the style features and replaces  the corresponding statistics in the content features with them. AdaIN performs effectively when both the style and content features are extracted using the same model.  However, in our framework, while the content features are rendered from the radiance field under the supervision of upsampled multi-level VGG features, the style features are directly extracted by VGG. This results in a pattern discrepancy between the content features and the style features, as shown in Fig.~\ref{fig:feature comparison}.

\begin{figure}[t!]
    \centering
    \includegraphics[width=\linewidth]{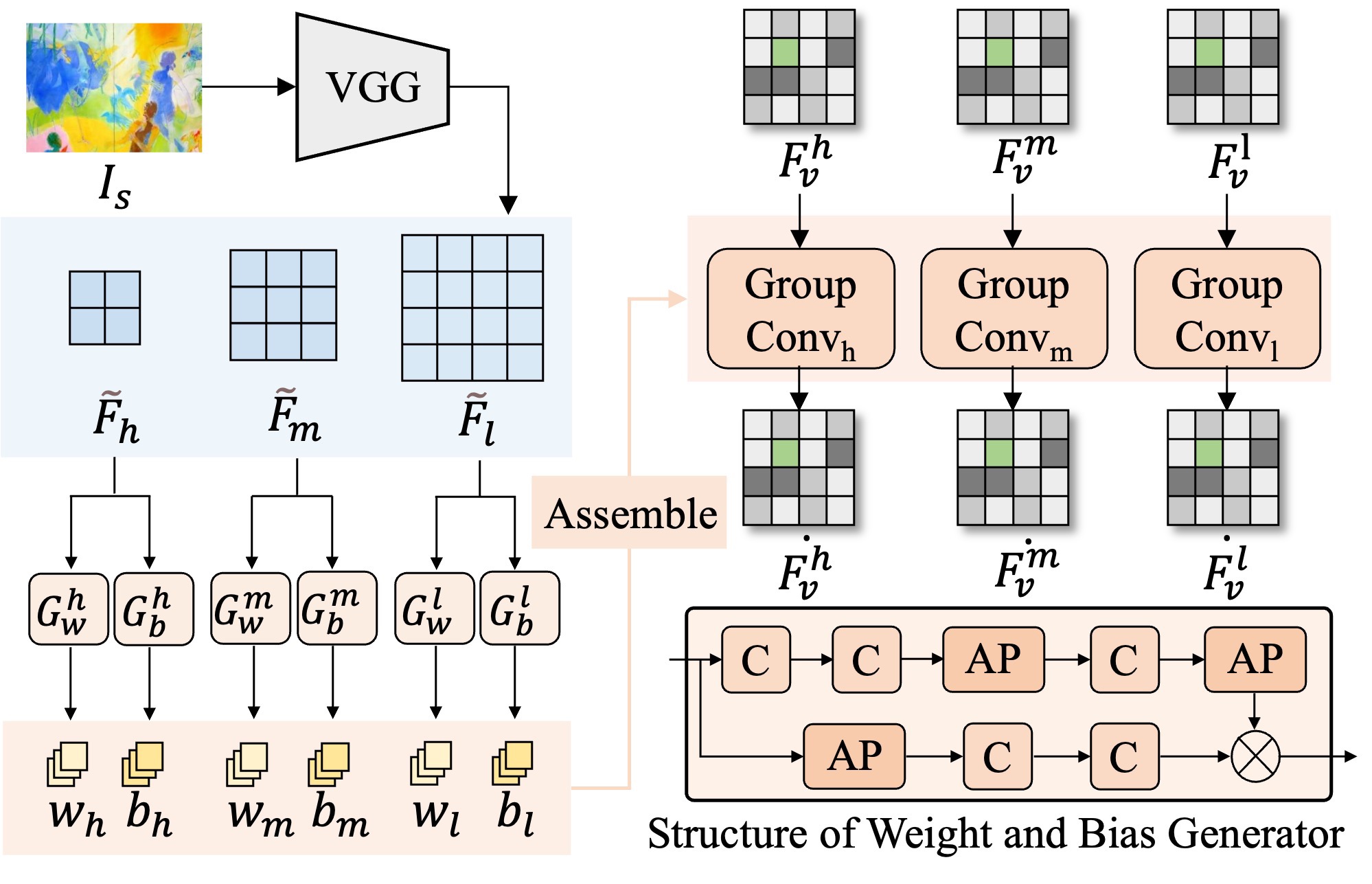}
    \caption{\textbf{Dynamic style injection (DSI).} DSI consists of two parts: (i) generating a set of weights and biases from the extracted style features, and (ii) applying group convolution to the rendered features using  the generated parameters. The structure of the weight and bias generator is shown at the bottom right, where the upper branch extracts  spatial information from the style features, and the lower branch extracts channel information. C denotes a 2D convolution layer, while AP refers to an adaptive average pooling layer.}
    \label{fig:DSI}
\end{figure}

To address the discrepancy between the rendered content features and the VGG extracted style features, it is essential for the injection process to effectively discern which style information should be integrated into the content features. Inspired by DIN, we propose a dynamic style injection mechanism that is in accordance with the requirement. The proposed mechanism uses convolution operators generated based on the style reference $I_s$ to adaptively modify the content features $F_v^{{\ell}}$. As shown in Fig.~\ref{fig:DSI}, we firstly use the VGG encoder to extract the style features at the same level, denoted  $\tilde{F}_{{\ell}}, {\ell}\in \left \{ {{\ell}}_{l},{{\ell}}_{m},{{\ell}}_{h} \right \}$. For each feature level ${\ell}$, exclusive weight and bias generators, represented as $\text{G}_{{\ell}}^w()$ and $\text{G}_{{\ell}}^b()$, are used to produce a set of  weights $w_{{\ell}} \in \mathbb{R}^{C_{{\ell}}}$ and biases $b_{{\ell}} \in \mathbb{R}^{C_{{\ell}}}$ based on their corresponding input features $\tilde{F}_{{\ell}}$. Here, $C_{{\ell}}$ is the number of feature channels in level ${\ell}$. The generated weights $w_{{\ell}}$ and biases $b_{{\ell}}$ are assembled into a group convolution operator $\text{GC}_{{\ell}}()$ that  acts on the content feature $F_v^{{\ell}} \in \mathbb{R}^{H_{{\ell}}\times W_{{\ell}} \times C_{{\ell}}}$. Mathematically, dynamic style injection can be written as:
\begin{equation}
\begin{aligned}
\dot{F}_v^{\ell} &= \text{GC}_{\ell}(F_v^{\ell} \mid w_\ell, b_\ell), \\
w_\ell &= \text{G}^{w}_\ell(\tilde{F}_{\ell}), b_\ell = \text{G}^{b}_\ell(\tilde{F}_{\ell}), \ell \in \{\ell_l, \ell_m, \ell_h\}.
\end{aligned}
\end{equation}
where $\dot{F}_{v}^{{\ell}}$ represents the injected features of level ${\ell}$ from view $v$. $w_{{\ell}}$ and $b_{{\ell}}$ denote the generated weight and bias parameters, respectively.

The weight and bias generators can be seen as processes that alter the rendered features according to the given style reference through the generated parameters, thereby achieving the aforementioned purpose in a learnable manner. Since the style descriptions in the reference are complicated, we design the generators to consider both spatial and channel information of the style reference. The spatial information is extracted through a sequence of convolutions followed by  adaptive average pooling operations, with the final pooling layer adaptively resizing the generated parameters to $1 \times 1$ dimensions. The channel weights are calculated using the same block as in SENet \cite{hu2019senet}, as illustrated by the lower branch in  Fig.~\ref{fig:DSI} (bottom right).

Finally, we use MLCD to gradually recover the stylized result $\tilde{I}_v$,  $\tilde{I}_v = \text{MLCD}(\dot{F}_v^{{{\ell}}_{l}},\dot{F}_v^{{{\ell}}_m},\dot{F}_v^{{{\ell}}_h}) \in \mathbb{R}^{H\times W\times 3}$, from the multi-level stylization features. During the stylization training stage, the common style and content loss from AdaIN is used:
\begin{equation}
    L_{cs} = L_c +\lambda L_s,
    \label{Eq:lcs}
\end{equation}
where $\lambda$ is a weight controlling stylization strength. In each iteration of the stylization training stage, a random style reference image is used to train the LIN, DSI and MLCD modules. During this process, LIN learns to suppress the styles in the content features. DSI learns to incorporate appropriate stylistic characteristics into the rendered features based on the style reference, while MLCD is trained jointly to decode these features into a stylized image. Once this stage is complete, MDS-NeRF has zero-shot style transfer capability.

While  style injection is based on  rendered 2D features, the operation can be converted into  3D space. Specifically, the pixel $\dot{f}_{v}^{{\ell}}$ in the injected feature map $\dot{F}_{v}^{{\ell}}$ can be written as:
\begin{equation}
    \dot{f}_{v}^{{\ell}}=f_{v}^{{\ell}}\otimes w_{{\ell}}+b_{{\ell}},
    \label{Eq:frldot}
\end{equation}
By combining Eq.~(\ref{Eq:frl}) with Eq.~(\ref{Eq:frldot}), we obtain:
\begin{equation}
\begin{split}
\dot{f}_{v}^{{\ell}}=\sum_{i=1}^{N}\omega_{i}\left(\bar{f}_{i}^{{\ell}}\otimes w_{{\ell}}+\frac{b_{{\ell}}}{\omega_{i}}\right), {{\ell}}\in\{{{\ell}}_{l},{{\ell}}_{m},{{\ell}}_{h}\},
\label{Eq:fvldot}
\end{split}
\end{equation}
where $\bar{f}_i^{{\ell}}$ is the queried point feature on corresponding ray $r$ at feature level ${\ell}$, and $\omega_i$ is the weight of point $i$. Through Eq.~(\ref{Eq:fvldot}), it is evident that dynamic style injection is equivalent to operating on the 3D point feature, thereby ensuring  multi-view consistent stylization.

\subsection{Style Transfer with 3D Reference}\label{sub:sec3.3}
\subsubsection{Rationale}
Existing methods generally transfer styles from 2D reference images. With the rapid increase in 3D artistic representations due to advances in 3D reconstruction technologies, as well as the emergence of 3D generation techniques that utilize text or single-view images as prompts, it has become essential to adapt stylization models for omni-view 3D-to-3D style transfer. In such circumstances, we have designed our MDS-NeRF framework to facilitate both 2D-to-3D and 3D-to-3D stylization. It should be noted that some optimization-based stylization methods like SNeRF and ARF can also achieve 3D-to-3D stylization. In contrast, our proposed MDS-NeRF enables zero-shot stylization without necessitating additional optimization of the content radiance fields. We present our approach to 3D-to-3D stylization in two scenarios: one employing existing 3D content as references and another utilizing 3D content derived from 2D images with assistance from generative models.

\subsubsection{Style transfer from 3D objects}\label{ST3D}
Here, we present our strategy for transferring styles from 3D content to accomplish omni-view stylization. The 3D content may be given in various forms, including multi-view images as well as implicit or explicit 3D representations. When the references are provided as multi-view images, their quantity may be limited, which can hinder the ability to satisfy requirements for stylization from any viewpoint. In such cases, we first train a 3D NeRF representation using the available views. Subsequently, we define the front view for both the style object and content object to align their poses. During the process of stylization, we synchronously alter their observation viewpoints. For each viewpoint, the rendered image of the style object serves as a 2D style reference and is sent to the DSI module to stylize the content from the current view. By repeating the process, we can obtain  stylized content results that exhibit a consistent stylistic pattern with respect to the style object across the viewpoint space.

Unrelated background areas in the style views can influence  stylization. To mitigate this influence, during the style injection stage, we render the mask of the style object and use the downsampled version of this mask to obscure irrelevant regions within the style features. This operation effectively restricts the weight and bias generators to focus on the subject features. However, it also results in dim and dark stylized results. This phenomenon occurs due to a significant reduction in the activation values of the style features, which subsequently affects the adaptive average pooling operation within the weight and bias generators.  To address this issue, we calculate an amplified scale based on the ratio of all pixels to the uncovered pixels in the rendered mask. We then multiply it by the extracted style features before sending them for  generation of weights and biases.

\subsubsection{Style transfer from lifted 3D objects}
For objects depicted in artistic paintings, we leverage generative models to derive their 3D representations. To mitigate the influence of the background during 3D generation and subsequent stylization, we employ the segmentation tool in \cite{sam}  to isolate an object from the painting prior to its use for 3D generation. As done for style transfer from 3D objects, we also adopt  multi-view images to represent the 3D style object. By referencing the isolated object image, we use generative models to synthesize multi-view images of the object. The remaining steps follow those described in the process of  style transfer from 3D objects.

\begin{figure*}[tp!]
   \centering
   \includegraphics[width=\linewidth]{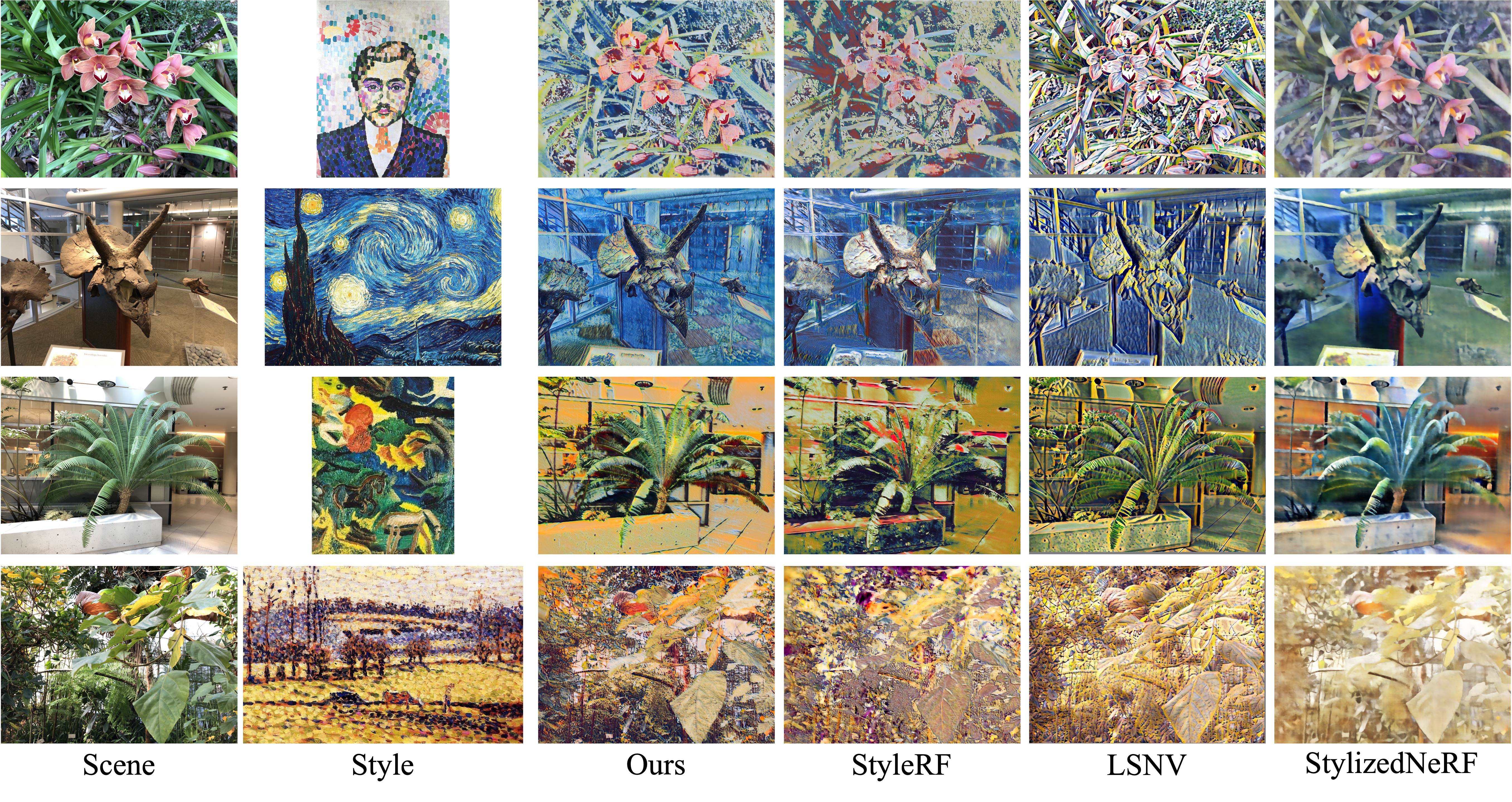}
   \caption{\textbf{Qualitative comparison.} We present 2D-to-3D style transfer results obtained by LSNV, StyleRF, StylizedNeRF and our proposed MDS-NeRF. Using the proposed multi-level feature grid and dynamic style injection module, MDS-NeRF produces results that preserve multi-level scene structures while effectively incorporating rich style characteristics.}
   \label{fig:qualitative compare1}
\end{figure*}

\begin{figure*}[tp!]
   \centering
   \includegraphics[width=\linewidth]{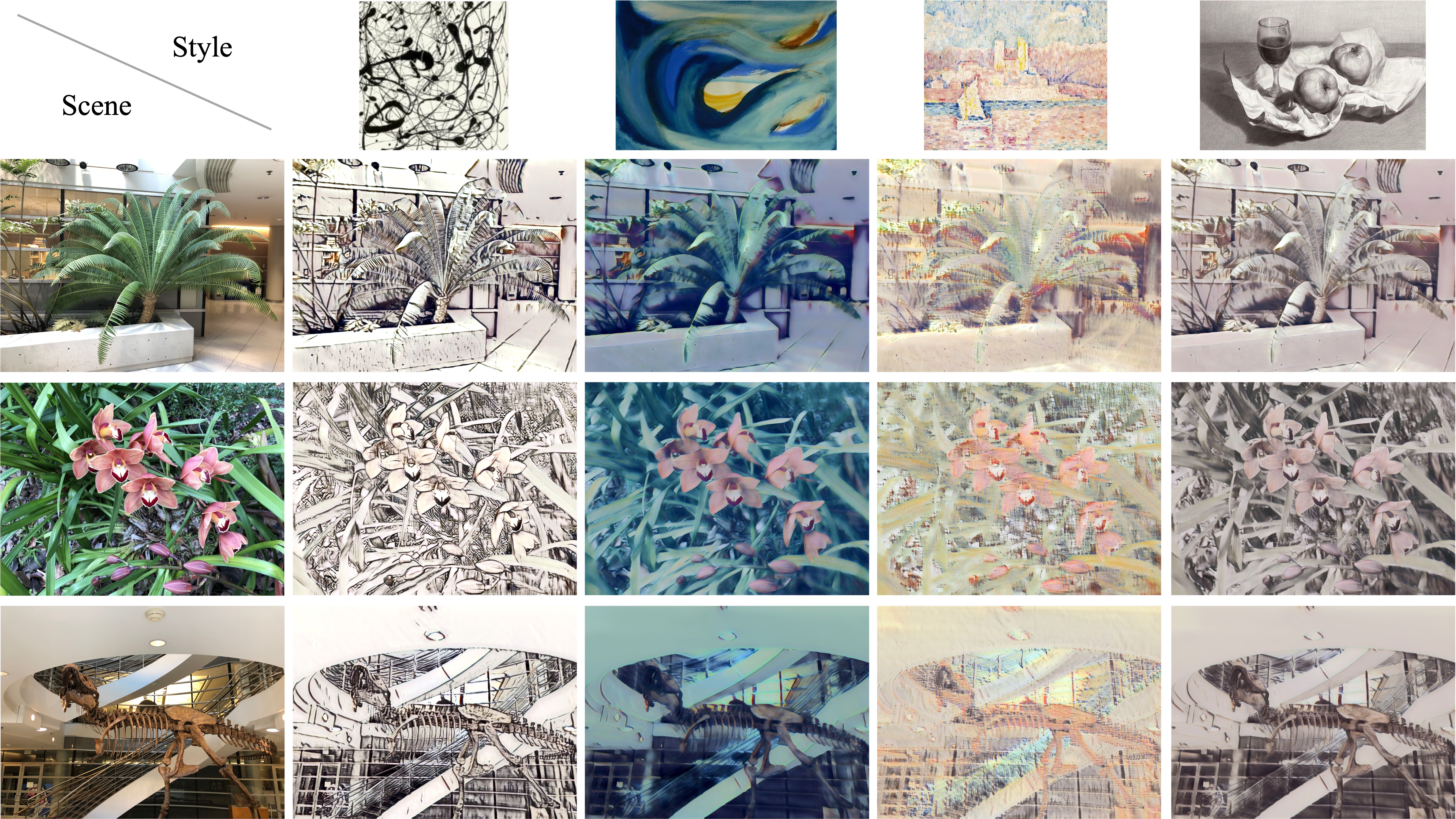}
   \caption{\textbf{2D-to-3D style transfer results.} We test our model's ability to transfer style using multiple reference styles with diverse artistic characteristics, including brush strokes, textures, and tonal variations. The results show that our model generates 3D scenes that faithfully incorporate these stylistic elements while preserving the structures of the 3D scenes.}
   \label{fig:qualitative result}
\end{figure*}

\section{Results and Evaluation}

\subsection{Experimental Setup}
\label{ExpSetup}
\subsubsection{Dataset and competing methods}
We chose the scenes from the NeRF synthetic and llff datasets \cite{10.1145/3306346.3322980}  as 3D content, training our proposed MDS-NeRF with the artistic images in the WikiArt dataset \cite{painter-by-numbers}, which was also used for stylization testing. We further used the synthetic scenes as the 3D style references and selected the object-centric paintings from the WikiArt dataset as 2D images to obtain the lifted 3D references as discussed in Sec.~\ref{sub:sec3.3}. Since  MDS-NeRF can achieve arbitrary style transfer, we compared our method to existing multi-style or arbitrary-style transfer methods, including StylizedNeRF, LSNV and StyleRF.

\subsubsection{Implementation}
To train the multi-level feature grid representation and the MLCD module, we performed a total of 30,000 iterations with a learning rate of $10^{-3}$. During the stylization training phase, we set  $\lambda$ in Eq.~(\ref{Eq:lcs}) to 30 and conducted 30,000 iterations to train the LINs and the DSI module with a learning rate of $10^{-3}$. Concurrently, we trained the MLCD with a learning rate of $10^{-5}$. We utilized the features from the ReLU1\_1, ReLU2\_1, and ReLU3\_1 layers of VGG19 as the low-, mid- and high-level style features. These multi-level style features were used for dynamic style injection and for computing the $L_f$ loss. Their channel dimensionalities, denoted as $C_{\mathrm{low}}, C_{\mathrm{mid}}$, and $C_{\mathrm{high}}$, were 64, 128, 256, respectively. For the lifted 3D-to-3D style transfer, we utilized Wonder3D, the image-to-3D method proposed in \cite{long2023wonder3d}, to generate the 3D object depicted in a painting as the lifted 3D style reference. Specifically, we reconstructed the multi-view images and normal maps of the object and then optimized a NeuS \cite{NeuS} representation for 6,000 iterations, resulting in the lifted 3D style reference.

For the competing methods, we trained StyleRF on the same dataset as our model, using their published hyperparameters. Following published guidelines, we trained StylizedNeRF directly on our test style images for efficient evaluation. For LSNV, we used the published checkpoint to obtain stylized results.

\subsection{Qualitative Evaluation}

\subsubsection{Comparison}
Fig.~\ref{fig:qualitative compare1} compares stylized results of StylizedNeRF, LSNV, StyleRF and our proposed MDS-NeRF on the llff dataset with 2D reference images. Visually, while StyleRF is capable of producing results with sufficient stylistic characteristics, it compromises edge structures, as StyleRF integrates only high-level stylistic features into the content. LSNV produces results with strong artistic styles; however, it fails to preserve the relative depth between content components due to an overly flattened stylization. For instance, in Fig.~\ref{fig:qualitative compare1} (row 4), the leaves and background appear indistinguishable. Although StylizedNeRF achieves a high level of content fidelity, the styles of its results diverge significantly from the reference images. As shown in Fig.~\ref{fig:qualitative compare1}, StylizedNeRF captures only the main color tones of the style references but fails to transfer stylistic details to the content. In contrast, our proposed MDS-NeRF effectively injects a rich variety of style characteristics (e.g., color, tone, brush strokes)  captured across multiple levels into the content. Consequently, it generates results that exhibit striking, globally harmonious stylized effects while preserving the spatial layouts and edge structures of the original scenes, as shown in Fig.~\ref{fig:qualitative compare1}.

\subsubsection{With 2D style reference}
As demonstrated in Fig.~\ref{fig:qualitative result}, our method effectively integrates diverse style characteristics into  3D scenes while preserving their structures. The input references encompass various artistic styles, including sketching, abstract art, impressionism, and classical still-life rendering, each introducing distinct stylistic attributes such as texture variations, line-based patterns, and painterly brush strokes. The results reflect the reference styles, demonstrating our method's robustness in transferring various styles. Additionally, the layouts and subjects remain clearly identifiable in the results, demonstrating MDS-NeRF's capability to effectively preserve scene structures.

\begin{figure*}[!tp]
    \centering
    \includegraphics[width=\linewidth]{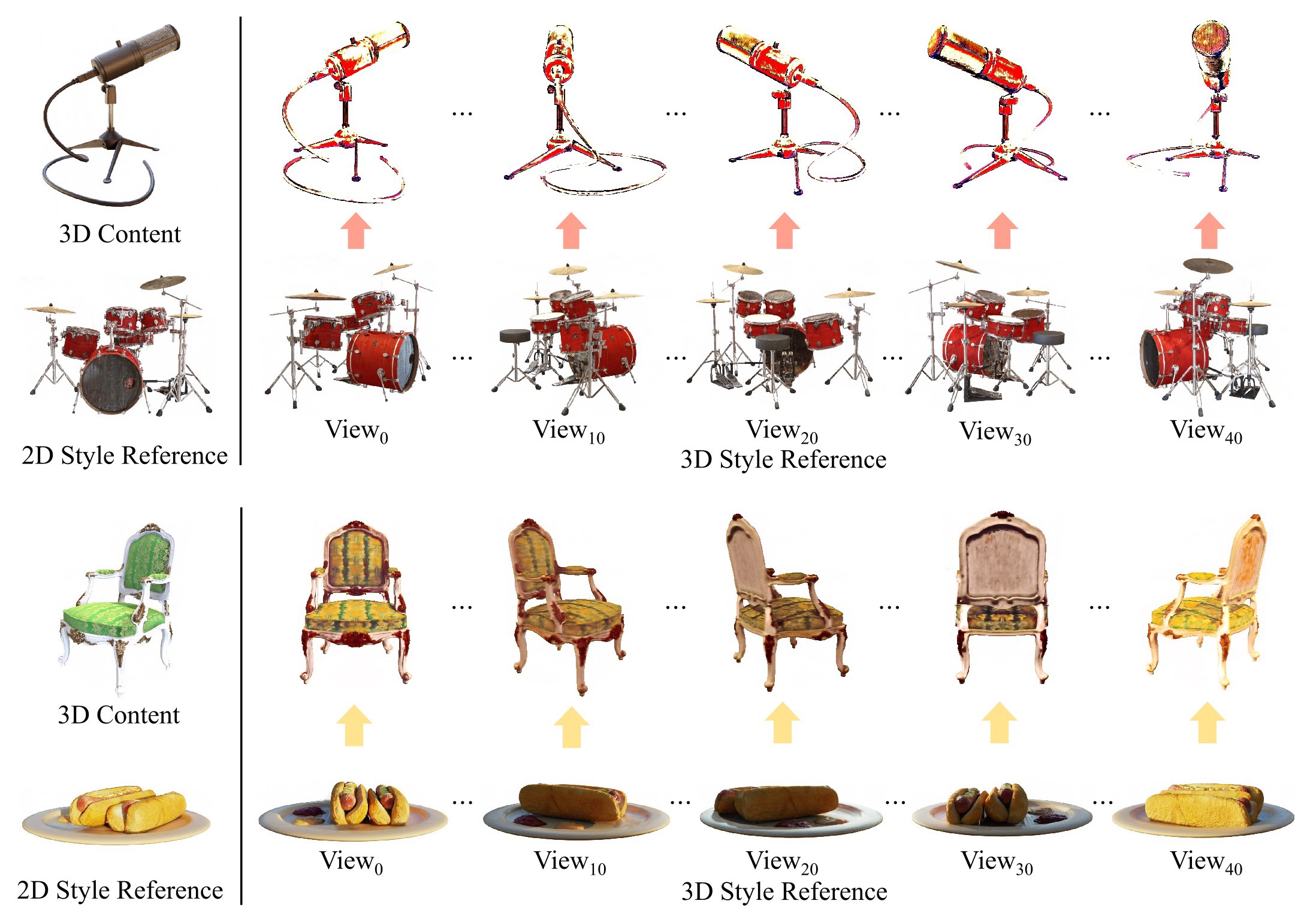}
    \caption{\textbf{3D-to-3D style transfer results.} We rendered a style image from each view and used it as the reference for stylizing the content scene for that view. As the observation view shifts within the scene, the viewpoint in the style field adjusts  accordingly, ensuring view-aligned 3D-to-3D style transfer.} 
    \label{fig:o2o_1}
\end{figure*}

\begin{figure*}[!tp]
    \centering
    \includegraphics[width=\linewidth]{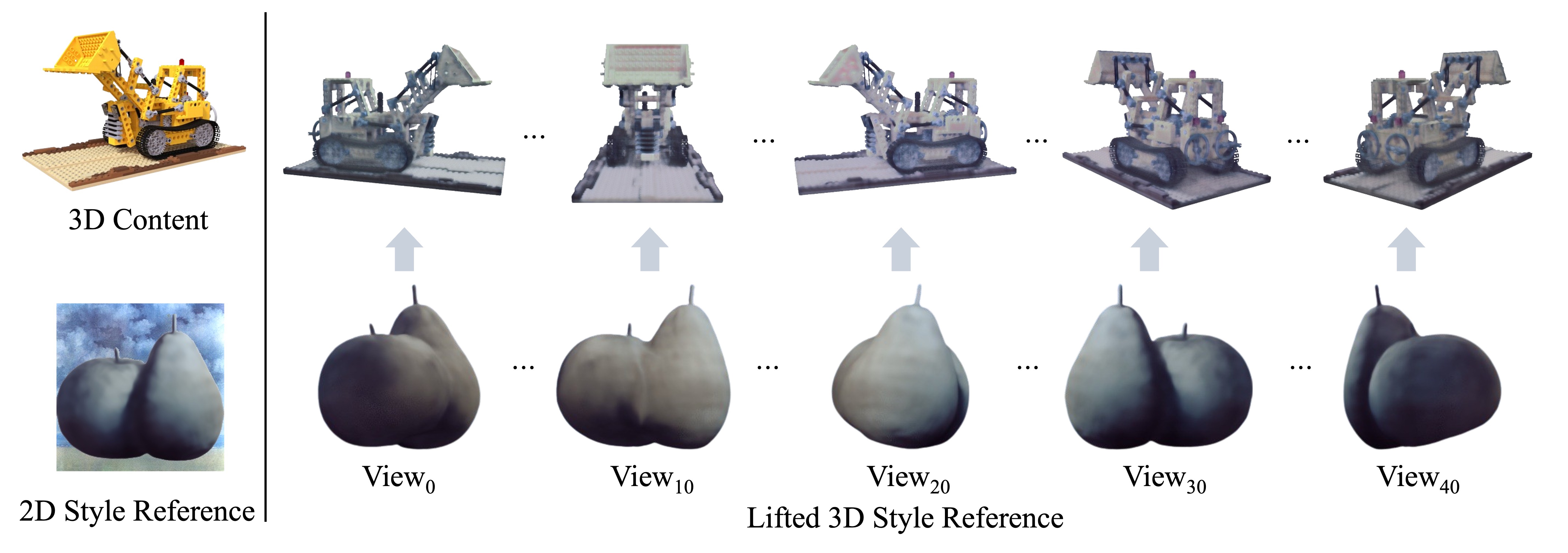}
    \caption{\textbf{Lifted 3D-to-3D style transfer results.} We adopted Wonder3D to reconstruct a 3D representation of the style object depicted in the 2D style reference.} 
    \label{fig:o2o_2}
\end{figure*}

\subsubsection{With 3D style reference}

We also evaluated our method for 3D-to-3D style transfer. Existing methods, such as StylizedNeRF, LSNV, and StyleRF, are used for 2D-to-3D stylization; therefore, we only present our results here. Fig.~\ref{fig:o2o_1} shows our  results with 3D references for which multi-view images are available. Fig.~\ref{fig:o2o_2} presents the results with a 3D style reference generated  from a given 2D artistic painting. As shown in Figs.~\ref{fig:o2o_1} and~\ref{fig:o2o_2},  the stylized results consistently exhibit style characteristics across all views. Particularly, the color and luminance in the stylized results change with the style fields as the view varies.
Note that the results of the lifted 3D-to-3D style transfer, shown in Fig.~\ref{fig:o2o_2}, are blurry. This is due to the blurry 3D reference, which is derived from a single-view image via image-to-3D generation. In contrast, style transfer from sharp views produces clear results, as shown in Figs.~\ref{fig:qualitative compare1}--\ref{fig:o2o_1}.

\begin{figure*}[!t]
   \centering
   \includegraphics[width=\linewidth]{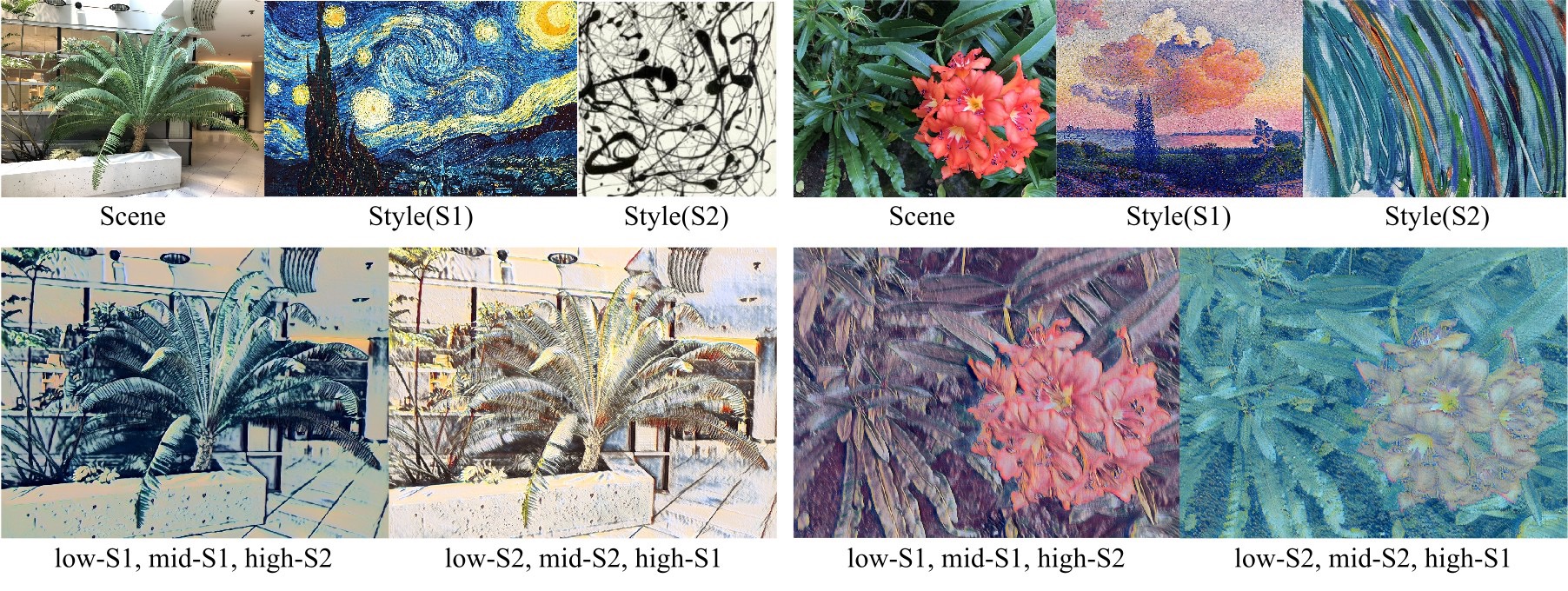}
   \caption{\textbf{Style Mixing.} The proposed multi-level feature grid representation enables  creative style mixing. By applying various style references at different levels of content features, we can obtain results with hybrid style characteristics.}
   \label{fig:style mix}
\end{figure*}

\begin{figure}[!t]
    \centering
    \includegraphics[width=\linewidth]{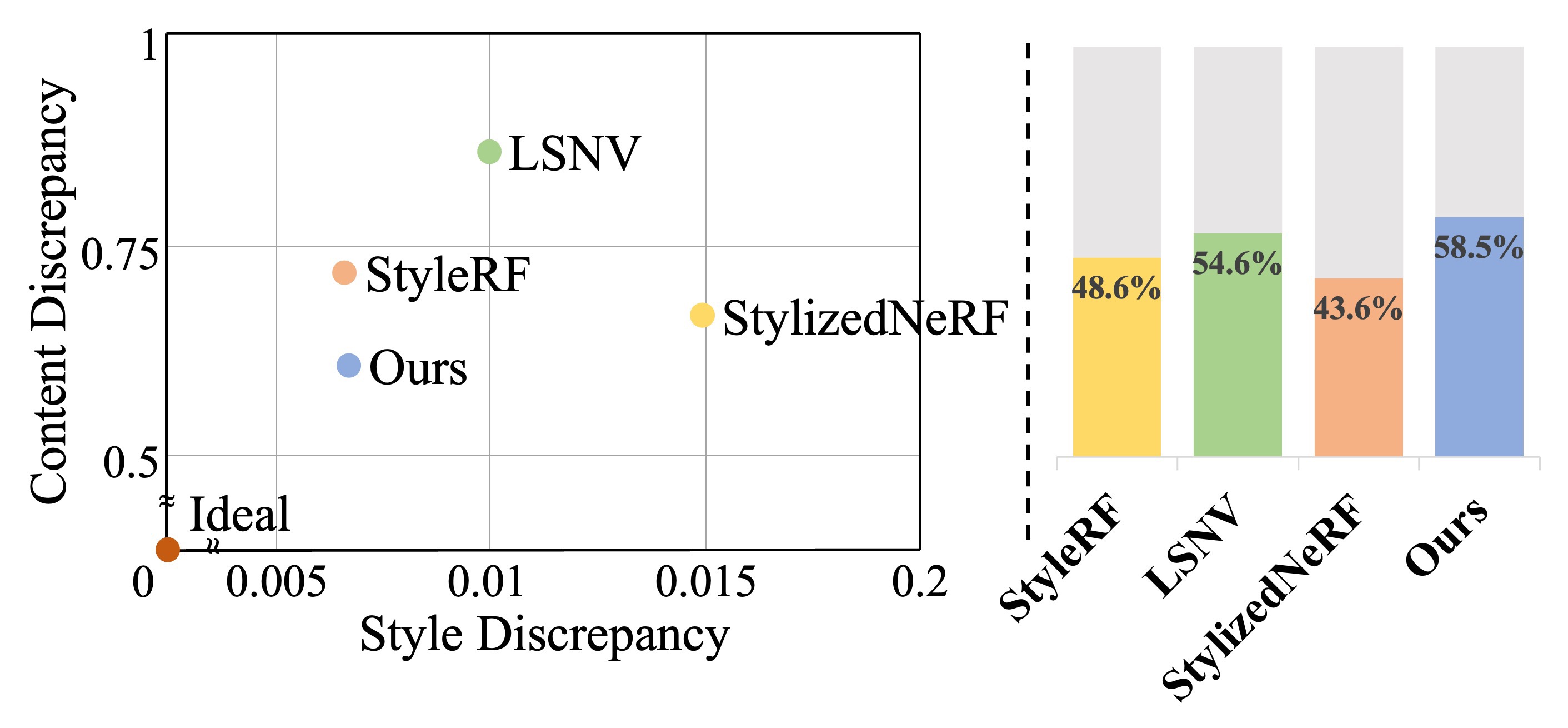}
    \caption{\textbf{Quantitative content/style evaluation and user study.}   We evaluated  content discrepancy using LPIPS distance based on the VGG model, and  style discrepancy using the style loss from AdaIN). In our user study, our questionnaire asked participants   to select their preferred stylized images from those generated by the  methods being compared. It should be noted that the total percentage from all methods exceeds 100\% because participants were allowed to choose multiple options.}
    \label{fig:userstudyandqualticomparison}
\end{figure}

\subsubsection{3D style mixing}
Thanks to the proposed multi-level feature grid representation, MDS-NeRF can transfer styles at a specific feature level and achieve flexible expression of style characteristics across different levels by combing  injected style features.
As shown in Fig.~\ref{fig:style mix}, we combined  style references S1 and S2 at different levels, and both stylized results effectively absorb the stylistic features of the corresponding levels and present them effectively. For instance, in the first example, the first result displays the texture pattern of S2, while showing the brightness and color style of S1. The second result shows the local textures of S1 along with the brightness and color of S2.

\subsection{Quantitative Evaluation}

\subsubsection{User study}
To objectively evaluate the results of our proposed MDS-NeRF, we conducted a user study based on the stylized results obtained by StylizedNeRF, LSNV, StyleRF, and our model using 2D style references. We used 6 scenes from the llff dataset and 30 randomly selected style references for testing. For each group of results, a total of 56 participants were asked to choose their preferred options in terms of content preservation and artistic stylization. Fig.~\ref{fig:userstudyandqualticomparison}  presents the user study results derived from the questionnaire, indicating that users preferred the results of LSNV and MDS-NeRF, mainly due to their superior stylistic effects. Notably, our MDS-NeRF emerged as the most preferred option, due to its effective stylization and high-quality content preservation.

\subsubsection{Content/Style evaluation}
We calculated the style loss from AdaIN based on the extracted VGG features as the style discrepancy indicators. We also calculated the LPIPS\_VGG \cite{Zhang_2018_CVPR} content loss as the content discrepancy indicator. Fig.~\ref{fig:userstudyandqualticomparison} shows the results. In comparison, MDS-NeRF performs well in stylized effects on account of its closer distance to the style reference. Although the style discrepancy of StyleRF is slightly smaller than that of MDS-NeRF, MDS-NeRF demonstrates superior performance in terms of content retention.

\begin{figure}[t!]
    \centering
    \includegraphics[width=\linewidth]{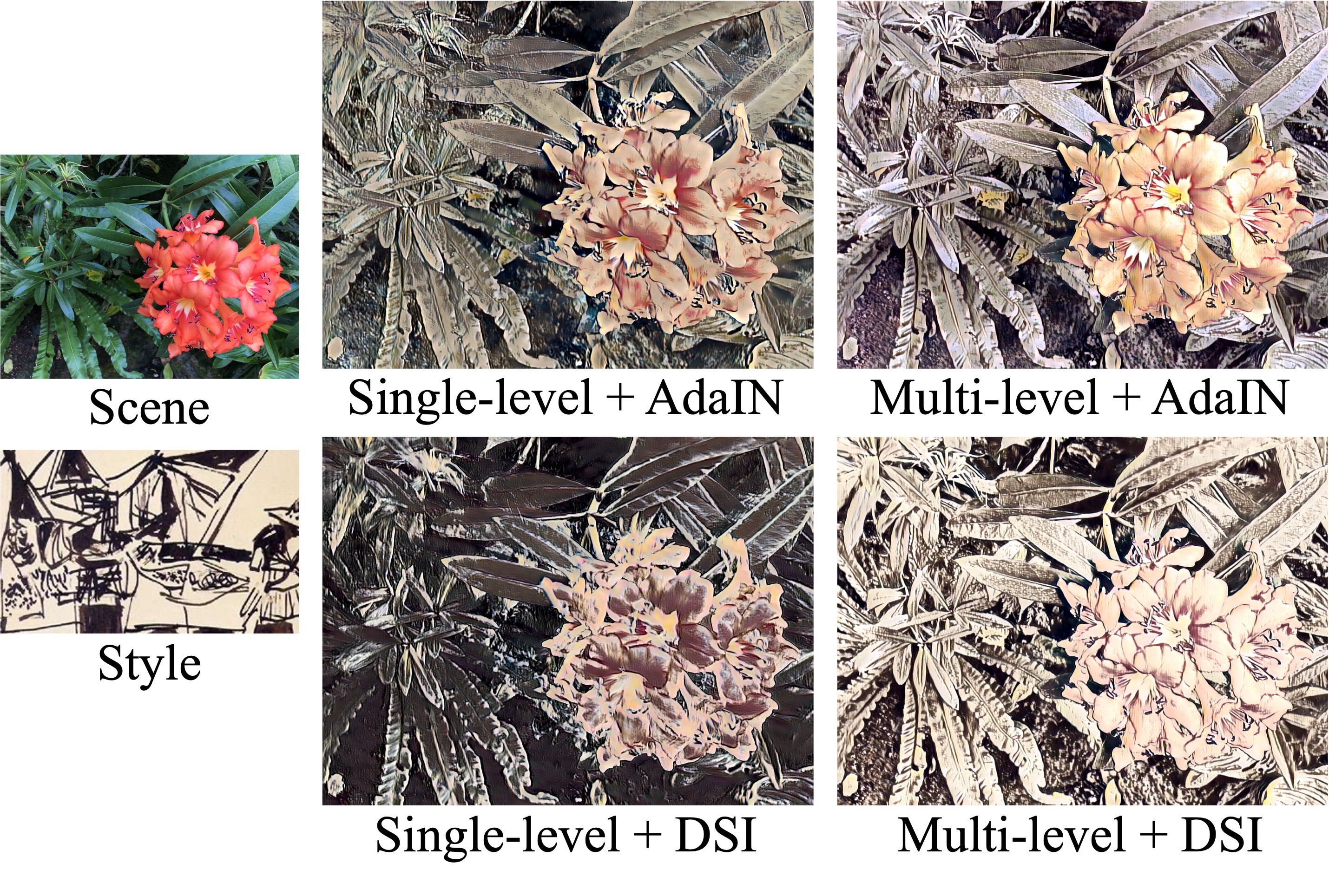}
    \caption{\textbf{Ablation study on the multi-level feature grid and DSI.} Single-level denotes the single-level feature grid used in StyleRF, while AdaIN denotes style injection based on feature statistics. }
    \label{fig:ablation study}
\end{figure}
\begin{figure*}[t!]
    \centering
    \includegraphics[width=\linewidth]{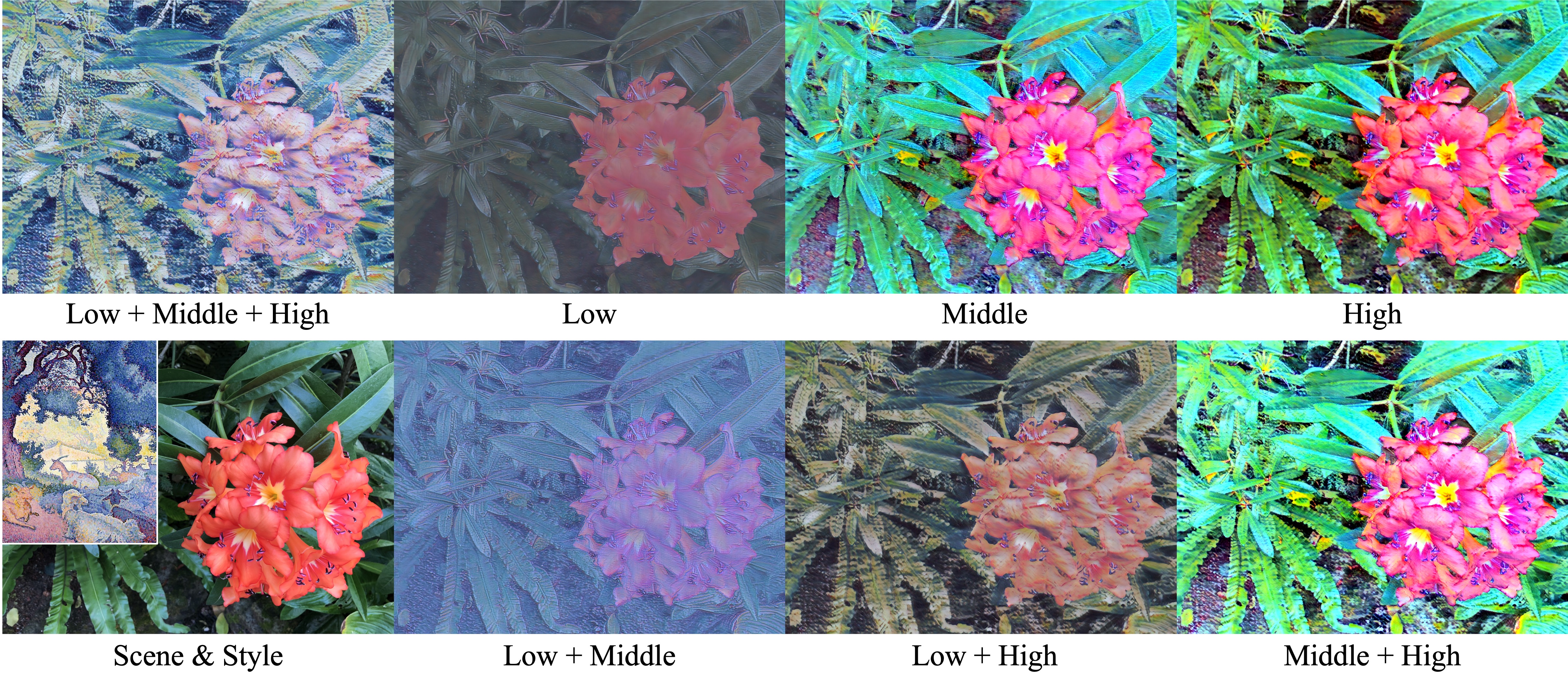}
    \caption{\textbf{Ablation study on style injection at different levels.} Low-level injection primarily incorporates basic style attributes, such as color and brightness, from the style reference, while high-level injection introduces fine-grained textures and abstract stylistic patterns. Mid-level injection produces styles that lie between those achieved through low- and high-level injections. By applying style transfer across all three levels, we obtained results with rich stylistic features while preserving the content structures.}
    \label{fig:ablation study2}
\end{figure*}
\subsection{Ablation Study}
Our proposed multi-level feature grid representation, in conjunction with the proposed adaptive dynamic style injection module, effectively facilitates 3D style transfer. To verify their effectiveness, we compared our method against three degenerate versions: the first utilized only the single high-level features along with the conventional AdaIN style injection method; the second employed the high-level features in conjunction with DSI; and the third used the multi-level feature grid alongside AdaIN. As shown in Fig.~\ref{fig:ablation study}, the approaches relying solely on the single high-level features failed to transfer color characteristics well. Furthermore, results obtained using AdaIN demonstrate inferior stylization effects compared to those generated via dynamic style injection. In contrast, our complete model exhibits superior content retention and enhanced multi-level stylized effects.

We also conducted an ablation study to analyze the impact of style injection at different feature levels within the DSI module. During the stylization process, DSI integrated specific levels of style features into the corresponding content features at the same levels. These DSI-processed features were then decoded into stylized images by the multi-level cascade decoder. As shown in Fig.~\ref{fig:ablation study2}, low-level injection primarily transfers basic style attributes, such as color and brightness, while high-level injection introduces fine-grained textures and abstract stylistic patterns. Mid-level injection produces stylistic effects that fall between those achieved through low- and high-level injections. Style injection across all levels effectively incorporates comprehensive style characteristics from the reference image while preserving the content structures.

\section{Conclusions}
The paper has presented a new zero-shot framework for 3D style transfer of NeRF, called MDS-NeRF. Once trained, MDS-NeRF can transfer styles from any reference. MDS-NeRF features a redesigned NeRF rendering pipeline that incorporates a multi-level feature grid representation and a multi-level cascade decoder. This design helps reveal the multi-scale structures of NeRF scenes, enabling rich style transfer while preserving spatial structures in the target scenes. Additionally, we proposed an adaptive dynamic style injection mechanism that selectively integrates preferred style characteristics into the multi-level feature grid, addressing discrepancies between content and style patterns. MDS-NeRF also supports style mixing and 3D-to-3D style transfer. Both qualitative and quantitative comparisons have demonstrated MDS-NeRF’s superior performance over existing multi-style and zero-shot style transfer methods.

Despite its advantages, MDS-NeRF cannot alter the shape of a content object to match a style object for tasks such as avatar stylization like NeRF-Art, as shown in Figs.~\ref{fig:o2o_1} and~\ref{fig:o2o_2}. Additionally, although the proposed 3D-to-3D style transfer approach achieves omni-view style transfer from 3D reference objects, it still has the following limitations. First, in the experiments of lifted 3D-to-3D style transfer, the quality of the results is affected by the low quality of the 3D references generated through current image-to-3D techniques. In particular, as shown in Fig.~\ref{fig:o2o_2}, the stylized objects appear blurry due to the blurred reference objects. This limitation can be mitigated by using more advanced image-to-3D generative models. Second, view-dependent stylization may disrupt  cross-view consistency of the original 3D content. In  future, we will focus on maintaining cross-view consistency while applying 3D-to-3D omni-view stylization within the framework of zero-shot radiance field style transfer. 

\section*{Availability of  data  and  materials}
The datasets used in this study are publicly available, and the data sources used are given in Section \ref{ExpSetup}.

\section*{Author contributions}
Conceptualization: Zesheng Li, Shuaibo Li; Methodology: Zesheng Li, Shuaibo Li, Wei Ma; Investigation---Zesheng Li; Writing---original draft preparation: Zesheng Li, Shuaibo Li; Writing---review and editing: Zesheng Li, Wei Ma, Jianwei Guo; Funding acquisition---Wei Ma, Jianwei Guo; Supervision: Hongbin Zha. All authors read and approved the final manuscript.

\section*{Acknowledgements}

This work was supported by the National Natural Science Foundation of China (62176010, 62262043, U22B2034), the Beijing Natural Science Foundation (4252029), the Open Project Program of State Key Laboratory of Virtual Reality Technology and Systems, Beihang University (VRLAB 2024B02), and Guangdong Basic and Applied Basic Research Foundation (2023B1515120026).

\section*{Declaration of competing interest}

The authors have no competing interests to declare that are relevant to the
content of this article.

\bibliographystyle{CVMbib}
\bibliography{refs}

\begin{thebibliography}{10}
\expandafter\ifx\csname urlstyle\endcsname\relax
  \providecommand{\doi}[1]{doi:\discretionary{}{}{}#1}\else
  \providecommand{\doi}{doi:\discretionary{}{}{}\begingroup \urlstyle{rm}\Url}\fi

\bibitem{3}
Jiang Y, Hedman P, Mildenhall B, Xu D, Barron JT, Wang Z, Xue T. {AligNeRF}: High-Fidelity Neural Radiance Fields via Alignment-Aware Training. In \emph{Proceedings of the IEEE/CVF Conference on Computer Vision and Pattern Recognition}, 2023, 46--55.

\bibitem{4}
Wang Q, Wang Z, Genova K, Srinivasan PP, Zhou H, Barron JT, Martin-Brualla R, Snavely N, Funkhouser T. {IBRNet}: Learning Multi-View Image-Based Rendering. In \emph{Proceedings of the IEEE/CVF Conference on Computer Vision and Pattern Recognition}, 2021, 4690--4699.

\bibitem{5}
Kerbl B, Kopanas G, Leimk{\"u}hler T, Drettakis G. {3D} Gaussian Splatting for Real-Time Radiance Field Rendering. \emph{ACM Transactions on Graphics}, 2023, 42(4): 139--1.

\bibitem{7}
Li Y, Fang C, Yang J, Wang Z, Lu X, Yang MH. Universal Style Transfer via Feature Transforms. \emph{Advances in Neural Information Processing Systems}, 2017: 385--395.

\bibitem{8}
Park DY, Lee KH. Arbitrary Style Transfer with Style-Attentional Networks. In \emph{proceedings of the IEEE/CVF Conference on Computer Vision and Pattern Recognition}, 2019, 5880--5888.

\bibitem{chen2021nips}
Chen H, Wang Z, Zhang H, Zuo Z, Li A, Xing W, Lu D, et~al.. Artistic Style Transfer with Internal-external Learning and Contrastive Learning. \emph{Advances in Neural Information Processing Systems}, 2021: 26561--26573.

\bibitem{11}
Liu S, Lin T, He D, Li F, Wang M, Li X, Sun Z, Li Q, Ding E. {AdaAttN}: Revisit Attention Mechanism in Arbitrary Neural Style Transfer. In \emph{Proceedings of the IEEE/CVF International Conference on Computer Vision}, 2021, 6649--6658.

\bibitem{12}
Li J, Xiang Y, Wu H, Yao S, Xu D. Optimal Transport-Based Patch Matching for Image Style Transfer. \emph{IEEE Transactions on Multimedia}, 2023, 25: 5927--5940.

\bibitem{fan2022unified}
Fan Z, Jiang Y, Wang P, Gong X, Xu D, Wang Z. Unified Implicit Neural Stylization. In \emph{In European Conference on Computer Vision}, 2022, 636--654.

\bibitem{Mu_2022_CVPR}
Mu F, Wang J, Wu Y, Li Y. {3D} {Photo Stylization}: Learning to Generate Stylized Novel Views from a Single Image. In \emph{Proceedings of the IEEE/CVF Conference on Computer Vision and Pattern Recognition}, 2022, 16273--16282.

\bibitem{UPST}
Chen Y, Yuan Q, Li Z, Liu Y, Wang W, Xie C, Wen X, Yu Q. {UPST-NeRF}: Universal Photorealistic Style Transfer of Neural Radiance Fields for {3D} Scene. \emph{IEEE Transactions on Visualization and Computer Graphics}, 2024: 1--13.

\bibitem{1}
Mildenhall B, Srinivasan PP, Tancik M, Barron JT, Ramamoorthi R, Ng R. {NeRF}: Representing Scenes as Neural Radiance Fields for View Synthesis. \emph{Communications of the ACM}, 2021, 65(1): 99--106.

\bibitem{NeRFArt}
Wang C, Jiang R, Chai M, He M, Chen D, Liao J. {NeRF-Art}: Text-Driven Neural Radiance Fields Stylization. \emph{IEEE Transactions on Visualization and Computer Graphics}, 2024, 30(8): 4983--4996.

\bibitem{IN2N}
Haque A, Tancik M, Efros AA, Holynski A, Kanazawa A. {Instruct-NeRF2NeRF}: Editing 3{D} Scenes with Instructions. In \emph{Proceedings of the IEEE/CVF International Conference on Computer Vision}, 2023, 19740--19750.

\bibitem{vica-nerf}
Dong J, Wang YX. {ViCA-NeRF}: View-Consistency-Aware {3D} Editing of Neural Radiance Fields. In \emph{Advances in Neural Information Processing Systems}, 2023, 61466--61477.

\bibitem{Huang_2022_CVPR}
Huang YH, He Y, Yuan YJ, Lai YK, Gao L. {StylizedNeRF}: Consistent {3D} Scene Stylization as Stylized NeRF via {2D-3D} Mutual Learning. In \emph{Proceedings of the IEEE/CVF Conference on Computer Vision and Pattern Recognition}, 2022, 18342--18352.

\bibitem{Liu_2023_CVPR}
Liu K, Zhan F, Chen Y, Zhang J, Yu Y, El~Saddik A, Lu S, Xing EP. {StyleRF}: Zero-shot {3D} Style Transfer of Neural Radiance Fields. In \emph{Proceedings of the IEEE/CVF Conference on Computer Vision and Pattern Recognition}, 2023, 8338--8348.

\bibitem{Chiang_2022_WACV}
Chiang PZ, Tsai MS, Tseng HY, Lai WS, Chiu WC. Stylizing {3D} Scene via Implicit Representation and HyperNetwork. In \emph{Proceedings of the IEEE/CVF Winter Conference on Applications of Computer Vision}, 2022, 1475--1484.

\bibitem{6}
Huang X, Belongie S. Arbitrary Style Transfer in Real-Time with Adaptive Instance Normalization. In \emph{Proceedings of the IEEE/CVF International Conference on Computer Vision}, 2017, 1501--1510.

\bibitem{simonyan2014very}
Simonyan K, Zisserman A. Very Deep Convolutional Networks for Large-Scale Image Recognition. \emph{ArXiv}, 2014, abs/1409.1556.

\bibitem{9}
Jing Y, Liu X, Ding Y, Wang X, Ding E, Song M, Wen S. Dynamic Instance Normalization for Arbitrary Style Transfer. In \emph{Proceedings of the AAAI Conference on Artificial Intelligence}, 2020, 4369--4376.

\bibitem{stylemesh}
H{\"o}llein L, Johnson J, Nie{\ss}ner M. {StyleMesh}: Style Transfer for Indoor {3D} Scene Reconstructions. In \emph{Proceedings of the IEEE/CVF Conference on Computer Vision and Pattern Recognition}, 2022, 6198--6208.

\bibitem{huang2021learning}
Huang HP, Tseng HY, Saini S, Singh M, Yang MH. Learning to Stylize Novel Views. In \emph{Proceedings of the IEEE/CVF International Conference on Computer Vision}, 2021, 13869--13878.

\bibitem{efficientnst}
Aurand J, Ortiz R, Nauer S, Azevedo VC. Efficient Neural Style Transfer for Volumetric Simulations. \emph{ACM Transactions on Graphics}, 2022, 41(6): 1--10.

\bibitem{intro_ARF}
Zhang K, Kolkin N, Bi S, Luan F, Xu Z, Shechtman E, Snavely N. {ARF}: Artistic Radiance Fields. In \emph{In European Conference on Computer Vision}, 2022, 717--733.

\bibitem{nguyen2022snerf}
Nguyen-Phuoc T, Liu F, Xiao L. {SNeRF}: Stylized Neural Implicit Representations for {3D} Scenes. \emph{ArXiv}, 2022, abs/2207.02363.

\bibitem{Ref-NPR}
Zhang Y, He Z, Xing J, Yao X, Jia J. Ref-NPR: Reference-Based Non-Photorealistic Radiance Fields for Controllable Scene Stylization. In \emph{Proceedings of the IEEE/CVF Conference on Computer Vision and Pattern Recognition}, 2023, 4242--4251.

\bibitem{GStyle}
Kov\'{a}cs AS, Hermosilla P, Raidou RG. G-Style: Stylized Gaussian Splatting. \emph{Computer Graphics Forum}, 2024, 43(7): e15259.

\bibitem{StylizedGS}
Zhang D, Yuan YJ, Chen Z, Zhang FL, He Z, Shan S, Gao L. StylizedGS: Controllable Stylization for 3D Gaussian Splatting. \emph{ArXiv}, 2024, abs/2404.05220.

\bibitem{Ha2016hyper}
Ha D, Dai AM, Le QV. HyperNetworks. \emph{ArXiv}, 2016, abs/1609.09106.

\bibitem{Rente2019TMM}
de~Oliveira~Rente P, Brites C, Ascenso J, Pereira F. Graph-Based Static {3D} Point Clouds Geometry Coding. \emph{IEEE Transactions on Multimedia}, 2019, 21(2): 284--299.

\bibitem{3dprinting}
Ezair B, Massarwi F, Elber G. Orientation analysis of {3D} objects toward minimal support volume in {3D}-printing. \emph{Computers \& Graphics}, 2015, 51: 117--124.

\bibitem{sun2022direct}
Sun C, Sun M, Chen HT. Direct Voxel Grid Optimization: Super-fast Convergence for Radiance Fields Reconstruction. In \emph{Proceedings of the IEEE/CVF Conference on Computer Vision and Pattern Recognition}, 2022, 5459--5469.

\bibitem{Fridovich-Keil_2022_CVPR}
Fridovich-Keil S, Yu A, Tancik M, Chen Q, Recht B, Kanazawa A. Plenoxels: Radiance Fields without Neural Networks. In \emph{Proceedings of the IEEE/CVF Conference on Computer Vision and Pattern Recognition}, 2022, 5501--5510.

\bibitem{Hou2024nerf}
Hou J, Zhang R, Wu Z, Meng W, Zhang X, Guo J. De-NeRF: Ultra-high-definition NeRF with deformable net alignment. \emph{Computer Animation and Virtual Worlds}, 2024, 35(3): e2240.

\bibitem{muller2022instant}
M{\"u}ller T, Evans A, Schied C, Keller A. Instant Neural Graphics Primitives with a Multiresolution Hash Encoding. \emph{ACM Transactions on Graphics}, 2022, 41(4): 1--15.

\bibitem{2}
Chen A, Xu Z, Geiger A, Yu J, Su H. {TensoRF}: Tensorial Radiance Fields. In \emph{In European Conference on Computer Vision}, 2022, 333--350.

\bibitem{Niemeyer_2021_CVPR}
Niemeyer M, Geiger A. {GIRAFFE}: Representing Scenes as Compositional Generative Neural Feature Fields. In \emph{Proceedings of the IEEE/CVF Conference on Computer Vision and Pattern Recognition}, 2021, 11453--11464.

\bibitem{NEURIPS2020_b4b75896}
Liu L, Gu J, Zaw~Lin K, Chua TS, Theobalt C. Neural Sparse Voxel Fields. In \emph{Advances in Neural Information Processing Systems}, 2020, 15651--15663.

\bibitem{unet}
Ronneberger O, Fischer P, Brox T. {U-Net}: Convolutional Networks for Biomedical Image Segmentation. In \emph{Medical Image Computing and Computer Assisted Intervention}, 2015, 234--241.

\bibitem{hu2019senet}
Hu J, Shen L, Sun G. Squeeze-and-Excitation Networks. In \emph{Proceedings of the IEEE/CVF Conference on Computer Vision and Pattern Recognition}, 2018, 7132--7141.

\bibitem{sam}
Kirillov A, Mintun E, Ravi N, Mao H, Rolland C, Gustafson L, Xiao T, Whitehead S, Berg AC, Lo WY, et~al.. Segment Anything. In \emph{Proceedings of the IEEE/CVF International Conference on Computer Vision}, 2023, 4015--4026.

\bibitem{10.1145/3306346.3322980}
Mildenhall B, Srinivasan PP, Ortiz-Cayon R, Kalantari NK, Ramamoorthi R, Ng R, Kar A. Local Light Field Fusion: Practical View Synthesis with Prescriptive Sampling Guidelines. \emph{ACM Transactions on Graphics}, 2019, 38(4): 1--14.

\bibitem{painter-by-numbers}
{Small yellow duck}, Kan W. Painter by Numbers. \url{https://kaggle.com/competitions/painter-by-numbers}, 2016, {K}aggle.

\bibitem{long2023wonder3d}
Long X, Guo YC, Lin C, Liu Y, Dou Z, Liu L, Ma Y, Zhang SH, Habermann M, Theobalt C, et~al.. Wonder3{D}: Single Image to {3D} using Cross-Domain Diffusion. In \emph{Proceedings of the IEEE/CVF Conference on Computer Vision and Pattern Recognition}, 2024, 9970--9980.

\bibitem{NeuS}
Wang P, Liu L, Liu Y, Theobalt C, Komura T, Wang W. NeuS: Learning Neural Implicit Surfaces by Volume Rendering for Multi-view Reconstruction. \emph{ArXiv}, 2021, abs/2106.10689.

\bibitem{Zhang_2018_CVPR}
Zhang R, Isola P, Efros AA, Shechtman E, Wang O. The Unreasonable Effectiveness of Deep Features as a Perceptual Metric. In \emph{Proceedings of the IEEE/CVF Conference on Computer Vision and Pattern Recognition}, 2018, 586--595.

\end{thebibliography}

\end{document}